%% file: main.tex
\documentclass[10pt,twocolumn,letterpaper]{article}

\usepackage{cvpr}              
\usepackage{multirow}
\input{preamble}

\definecolor{cvprblue}{rgb}{0.21,0.49,0.74}
\usepackage[pagebackref,breaklinks,colorlinks,allcolors=cvprblue]{hyperref}

\def\paperID{12849} 
\def\confName{CVPR}
\def\confYear{2025}

\title{SVDC: Consistent Direct Time-of-Flight Video Depth Completion with Frequency Selective Fusion}

\author{
Xuan Zhu$^{1}$,
~~Jijun Xiang$^{1}$,
~~Xianqi Wang$^{1}$,
~~Longliang Liu$^{1}$, \\
~~Yu Wang$^{2}$,
~~Hong Zhang$^{2}$,
~~Fei Guo$^{2}$,
~~Xin Yang$^{1}$\footnotemark[2]\\
[2mm]
$^1$~Huazhong University of Science and Technology \quad $^2$~Honor Device Co., Ltd \\
{\tt\small \{xuanzhu, jijunx, xianqiw, longliangl, xinyang2014\}@hust.edu.cn}
}

\begin{document}
\maketitle
\input{sec/0_abstract}    

\renewcommand{\thefootnote}{\fnsymbol{footnote}}
\footnotetext[2]{Corresponding author.}

\input{sec/1_intro}
\input{sec/2_formatting}
\input{sec/3_finalcopy}
\input{sec/4_Experiments}
\input{sec/5_Conclusion}
{
    \small
    \bibliographystyle{ieeenat_fullname}
    \bibliography{main}
}

\input{sec/X_suppl}

\end{document}

%% file: preamble.tex
\usepackage{float}

%% file: sec/0_abstract.tex
\begin{abstract}
Lightweight direct Time-of-Flight (dToF) sensors are ideal for 3D sensing on mobile devices. However, due to the manufacturing constraints of compact devices and the inherent physical principles of imaging, dToF depth maps are sparse and noisy. In this paper, we propose a novel video depth completion method, called SVDC, by fusing the sparse dToF data with the corresponding RGB guidance. Our method employs a multi-frame fusion scheme to mitigate the spatial ambiguity resulting from the sparse dToF imaging. Misalignment between consecutive frames during multi-frame fusion could cause blending between object edges and the background, which results in a loss of detail. To address this, we introduce an adaptive frequency selective fusion (AFSF) module, which automatically selects convolution kernel sizes to fuse multi-frame features. Our AFSF utilizes a channel-spatial enhancement attention (CSEA) module to enhance features and generates an attention map as fusion weights. The AFSF ensures edge detail recovery while suppressing high-frequency noise in smooth regions. To further enhance temporal consistency, We propose a cross-window consistency loss to ensure consistent predictions across different windows, effectively reducing flickering. Our proposed SVDC achieves optimal accuracy and consistency on the TartanAir and Dynamic Replica datasets. Code is
 available at \href{https://github.com/Lan1eve/SVDC}{\textcolor{magenta}{https://github.com/Lan1eve/SVDC}}.
\end{abstract} 

%% file: sec/1_intro.tex
\section{Introduction}
\label{sec:intro}

Obtaining consistent and accurate depth video on mobile devices is essential for constructing precise 3D scene models and plays a significant role in applications such as 3D reconstruction and AR/VR\cite{baruchARKitScenesDiverseRealWorld2022}. With the rapid advancement of sensor technology, novel lightweight direct Time-of-Flight (dToF) sensors\cite{ronchiniximenesModularDirectTimeFlight2019} have created new opportunities for depth enhancement research. By emitting laser pulses and measuring the reflection time, dToF sensors acquire depth information and offer advantages such as compact size, low cost, and energy efficiency. Consequently, they have attracted considerable attention from both academia and industry\cite{lindellSinglephoton3DImaging2018,morimotoMegapixelTimegatedSPAD2019}.
\begin{figure}[t]
  \centering
   \includegraphics[width=1\linewidth]{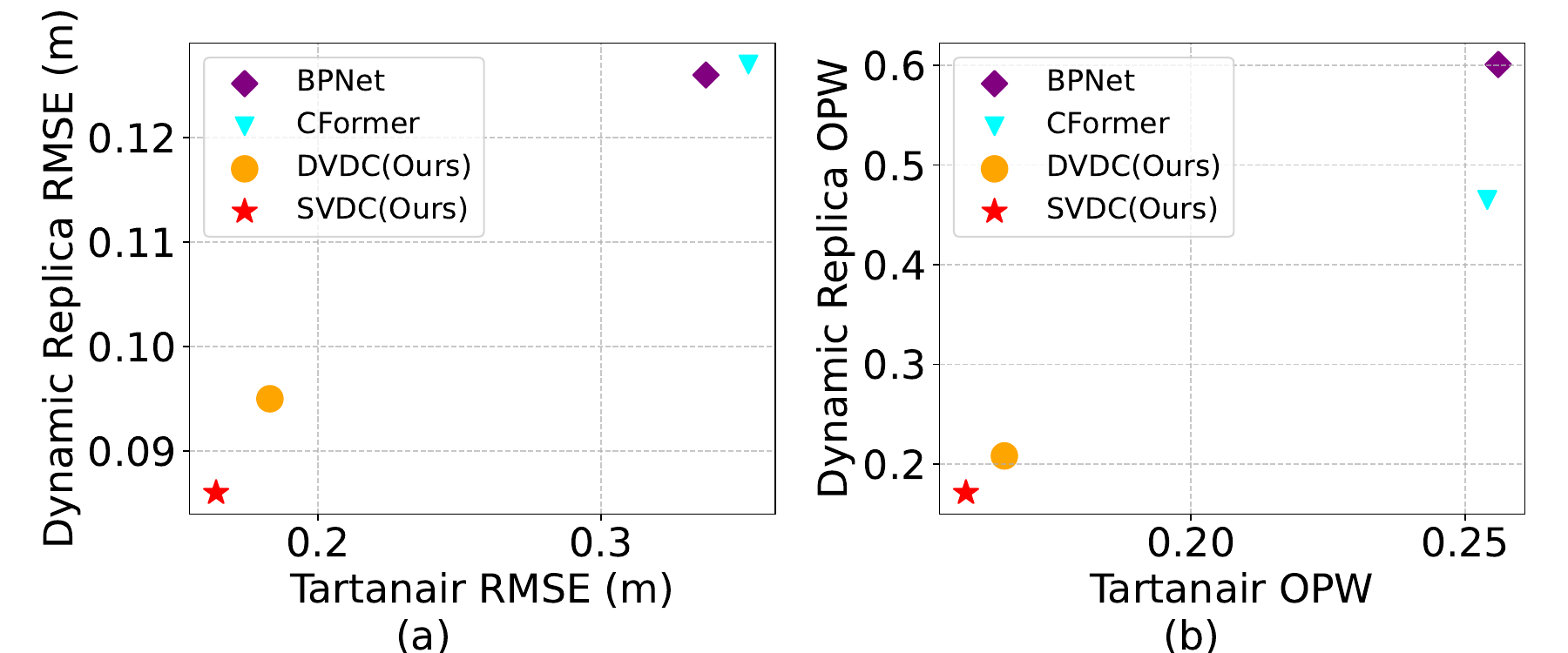}

   \caption{Comparisons with state-of-the-art (SOTA) methods on the TartanAir and Dynamic Replica datasets. \textbf{Left}: Accuracy metric RMSE$\downarrow$. \textbf{Right}: Temporal consistency metric OPW$\downarrow$\cite{wangLessMoreConsistent2022}. Our proposed approach achieves superior accuracy and consistency compared to per-frame depth completion methods.}
   \label{fig:Compare_with_sota_fg1}
\end{figure}

DToF sensors, depending on their type, typically return two forms of depth information: low-resolution depth maps or sparse depth maps. The low-resolution depth map provides detailed and accurate depth information for patch regions within the image, such as the mean and variance of depth values, and even histograms of the depth distribution. Previous research has focused on depth super-resolution tasks for dToF data\cite{sunConsistentDirectTimeFlight2023,liDELTARDepthEstimation2022}. Deltar\cite{liDELTARDepthEstimation2022} proposed a two-branch model in which one branch encodes RGB information while the other encodes the dToF low-resolution depth map. Features from these two branches are fused at various levels within the decoder, using RGB information to guide the recovery of the low-resolution depth map. DVSR\cite{sunConsistentDirectTimeFlight2023} introduced a video-based dToF depth super-resolution algorithm that leverages RGB information as guidance and incorporates an optical flow-guided deformable convolution module\cite{chanBasicVSRImprovingVideo2022} to aggregate and propagate multi-frame features. This approach allows multi-frame information to complement each other, ultimately predicting accurate and consistent high-resolution depth video. 

However, obtaining sparse depth maps from dToF sensors is more convenient and cost-effective than low-resolution depth maps. Consequently, lightweight dToF that provide sparse depth maps have gained widespread adoption in mobile devices. Unlike depth completion tasks based on automotive LiDAR, lightweight dToF in mobile devices can capture depth information for only a very small fraction of image pixels (e.g. $\sim20 \times 30$ for iPhone dToF), creating a significant sparsity challenge for the completion process. As a result, fusing sparse depth maps from dToF demands models with stronger inference capabilities and better adaptability.

dToF low-resolution depth maps return the mean depth value within image patches, resulting in minimal variations between different frames. In contrast, the sparse depth maps provided by dToF offer precise depth values of pixels, leading to more noticeable depth changes between frames. Simply using optical flow networks\cite{ranjanOpticalFlowEstimation2017,xuGMFlowLearningOptical2022} to align and fuse multi-frame features can easily result in feature misalignment due to inaccuracies in optical flow estimation, which in turn causes blending issues between object edges and the background. Furthermore, in the pursuit of temporal consistency in depth estimation, the greater variability of sparse depth maps poses a more significant temporal consistency challenge for the completion of dToF sparse depth maps. Existing video depth estimation methods\cite{luoConsistentVideoDepth2020,liPracticalConsistentVideo2023,wangNeuralVideoDepth2023,wangLessMoreConsistent2022} typically use a window-based approach for training and inference. Within each window, consecutive frame features are fused using optical flow alignment or cross-attention mechanisms, and temporal consistency losses are applied to enforce the stability of depth predictions. However, these methods often overlook consistency constraints across windows. Although the predictions within a window are consistent, there are noticeable differences between adjacent windows, resulting in flickering in the depth prediction results.

To address the issue of incorrect depth propagation due to feature misalignment during multi-frame fusion, which results in blending between object edges and the background, we propose an Adaptive Frequency Selective Fusion (AFSF) module. By adaptively selecting convolution kernel sizes based on frequency characteristics, the module mitigates the impact of optical flow misalignment that causes blending of objects and background.

To achieve adaptive frequency selection, we propose the Channel-Spatial Enhancement Attention (CSEA) module, which enhances high-frequency information in features while extracting an attention map to distinguish between different frequency areas. Through adaptive selection, the AFSF module applies smaller convolution kernels to misaligned object edges to preserve high-frequency details at object boundaries, mitigating the abnormal blending of object edges and background caused by misalignment. For smooth low-frequency regions, larger convolution kernels are used to suppress abnormal high-frequency noise interference in low-frequency areas.

Moreover, we propose a lightweight video depth completion model called DVDC. Based on this framework, we further integrate the CSEA and AFSF modules, leading to an enhanced model called SVDC.

In addition, we introduce a cross-window consistency loss to address the lack of consistency constraints across windows and to ensure consistent predictions. During training, each window contains three consecutive frames, and by incorporating SILoss\cite{eigenDepthMapPrediction2014}, we minimize the prediction differences for the same frames predicted by different windows, which enhances cross-window prediction consistency.

We evaluate our DVDC and SVDC on the TartanAir\cite{wangTartanAirDatasetPush2020} and Dynamic Replica\cite{karaevDynamicStereoConsistentDynamic2023} public datasets, demonstrating the effectiveness of each component through ablation studies. Compared to per-frame processing baselines, our multi-frame method significantly improves both prediction accuracy and temporal consistency. Our approach achieves state-of-the-art performance as shown in \cref{fig:Compare_with_sota_fg1}, while requiring the fewest parameters.

Our main contributions can be summarized as follows:
\begin{itemize}
    \item We propose a lightweight video depth completion model called DVDC that fuses multi-frame features to help the completion of sparse and noisy sparse dToF depth maps.
\end{itemize}

\begin{itemize}
    \item We introduce the CSEA and AFSF modules, which enhance feature representations, generate attention maps and adaptively fuse multi-frame features in different regions. By incorporating CSEA and AFSF modules into the DVDC model, we obtain the SVDC model.
\end{itemize}

\begin{itemize}
    \item We propose a cross-window temporal consistency loss, which effectively improves the temporal consistency of the predicted results.
\end{itemize}

\begin{itemize}
    \item Our model outperforms existing depth completion approaches, achieving superior accuracy and consistency.
\end{itemize}

\begin{figure*}[t]
  \centering
  \includegraphics[width=1.0\linewidth]{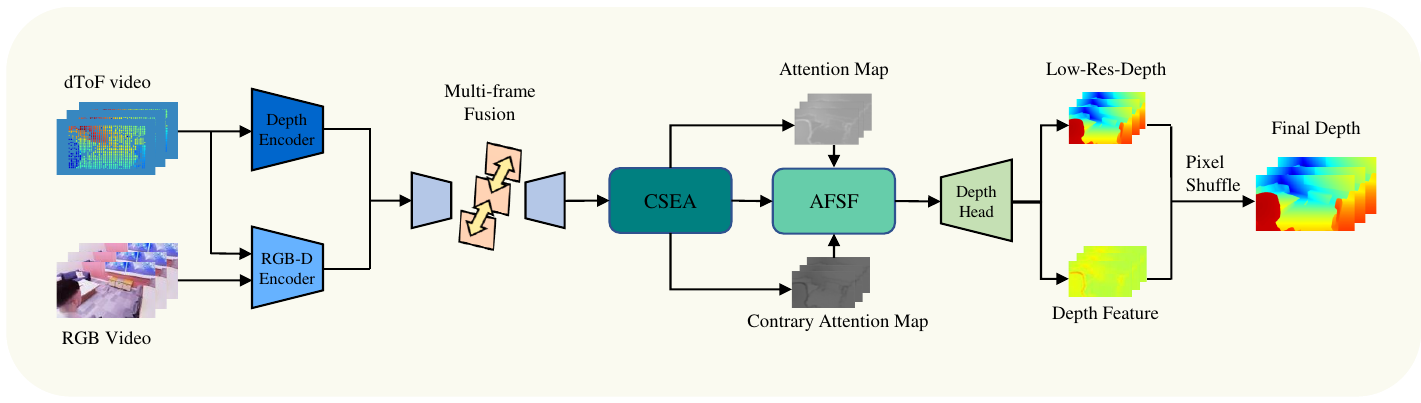}

  \caption{Overview of the proposed SVDC network. The CSEA module enhances multi-frame features and extracts attention maps to guide the AFSF module in selectively fusing multi-frame features. Finally, the low-resolution depth is obtained through the depth head and refined using the feature-guided pixel shuffle module to produce the final depth.}
  \label{fig:backbone_fig2}
\end{figure*}

%% file: sec/2_formatting.tex
\section{Related Work}
\label{sec:formatting}

\textbf{Depth enhancement.} Depth enhancement methods aim to restore degraded depth maps to high-quality ones. Generally, these methods are categorized into two main approaches: depth completion\cite{huPENetPreciseEfficient2021,wongUnsupervisedDepthCompletion2021,zhangCompletionFormerDepthCompletion2023,tangBilateralPropagationNetwork2024,wangLRRULongshortRange2023} and depth super-resolution\cite{heFastAccurateRealWorld2021,metzgerGuidedDepthSuperResolution2023,yuanRecurrentStructureAttention2023}. Most depth completion methods rely on sparse depth maps obtained from LiDAR and typically follow a two-step process\cite{huPENetPreciseEfficient2021,zhangCompletionFormerDepthCompletion2023}: first, fusing color and depth information, and then applying post-processing\cite{chengCSPNLearningContext2020,parkNonlocalSpatialPropagation2020}. In some approaches, sparse depth maps are preprocessed before fusion to improve performance. By integrating sparse depth maps with RGB images and iteratively refining the depth estimates during post-processing, these methods leverage sparse depth data to alleviate over-smoothing issues.

In comparison, depth super-resolution methods aim to upscale a low-resolution depth map to a higher one, often through color-guided progressive upsampling or by modeling the depth super-resolution task as a pixel-to-pixel mapping\cite{lutioGuidedSuperResolutionPixelPixel2019}. However, both LiDAR-based depth completion and depth super-resolution tasks differ significantly from depth enhancement tasks that utilize degraded depth maps of dToF sensor. dToF data is typically much sparser and relies on the principles of physical imaging, introducing considerable noise. Therefore, directly applying existing methods is insufficient to address the specific challenges of dToF data.

Some studies have focused on depth enhancement for dToF data. Deltar\cite{liDELTARDepthEstimation2022} introduces an attention mechanism between patch blocks and RGB pixels to guide the restoration of low-resolution dToF depth maps. DVSR\cite{sunConsistentDirectTimeFlight2023} addresses low-resolution dToF video streams by using optical flow and deformable convolutions to fuse and propagate information across frames, further improving accuracy and consistency. In the depth completion task for sparse dToF depth map, EMDC\cite{houLearningEfficientMultimodal2023} employs a two-branch network and designs an FCSPN network with a large receptive field to adapt to the distribution characteristics of sparse dToF depth map. This network iteratively refines depth estimations, achieving promising results.

It is worth noting that existing sparse dToF depth map completion tasks are primarily based on single-frame data, whereas our objective is to achieve sparse depth point completion for dToF in video streams. In video sequences, compared to the low-resolution depth maps that return the mean depth value of patch blocks, the sparse dToF depth maps exhibit much greater variation over time. Simply using multi-frame fusion networks can easily cause incorrect depth propagation due to feature misalignment, resulting in blending issues between object edges and the background.

\textbf{Video depth estimation.} In mobile devices, the input data for depth estimation is mostly in the form of the video stream, providing multi-view and temporal information. This places a higher demand on temporal consistency. Current video depth estimation approaches aim to achieve temporal consistency and can be categorized into two main types: test-time training (TTT) methods\cite{luoConsistentVideoDepth2020,kopfRobustConsistentVideo2021,zhangConsistentDepthMoving2021} and learning-based methods. TTT methods, like CVD\cite{luoConsistentVideoDepth2020}, use pre-trained monocular depth estimation models fine-tuned with geometric constraints and camera poses. This approach enhances accuracy but comes with high computational costs and struggles in occluded or textureless regions.

Learning-based methods can be further divided into two categories. One category integrates temporal information within deep learning networks, training depth models directly with spatial and temporal supervision. For instance, TCMonoDepth\cite{liEnforcingTemporalConsistency2021} introduces temporal consistency loss for depth estimation, ST-CLSTM\cite{zhangExploitingTemporalConsistency2019} models temporal relationships by incorporating LSTM\cite{shiConvolutionalLSTMNetwork2015}, and FMNet\cite{wangLessMoreConsistent2022} combines convolutional self-attention to recover depth for masked frames from unmasked frames. VITA\cite{xianViTAVideoTransformer2024} employs Transformer with temporal embeddings in the attention blocks, while MAMo\cite{yasarlaMAMoLeveragingMemory2023} introduces memory update and memory attention mechanisms to leverage temporal information. These methods effectively reduce depth flickering between frames but are time-consuming. Another category utilizes post-processing techniques\cite{wangNeuralVideoDepth2023,liPracticalConsistentVideo2023}, where the predictions from pre-trained monocular depth estimation models\cite{ranftlVisionTransformersDense2021,ranftlRobustMonocularDepth2022} are fed into a stabilizer network to further enhance the consistency of the model's predictions. However, due to memory limitations, most existing methods perform training based on windows, while overlooking information across different windows. This often results in noticeable flickering in the prediction between adjacent windows. We improve cross-window consistency by leveraging a cross-window temporal consistency loss.

\begin{figure*}[!htbp]
  \centering
  \includegraphics[width=1.0\linewidth]{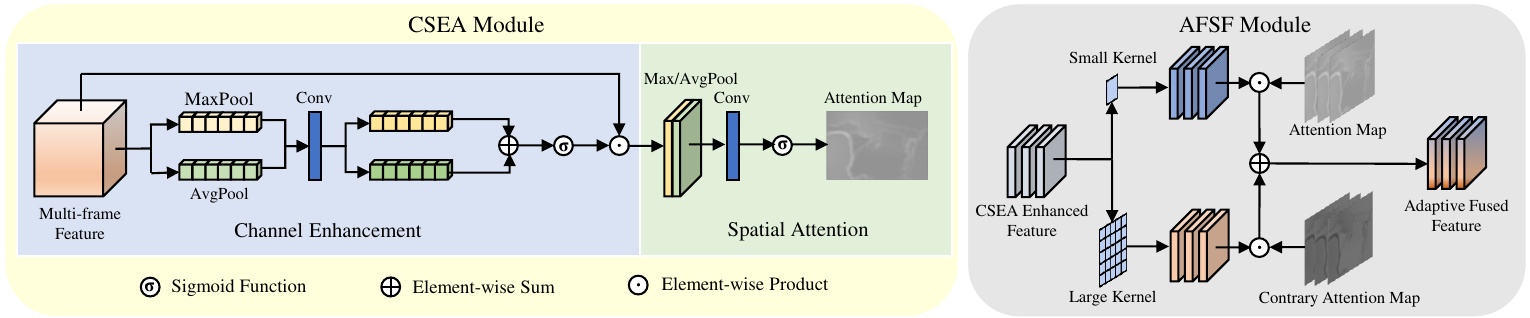}

  \caption{The proposed CSEA and AFSF architectures. Left: CSEA module. Right: AFSF module.}
  \label{fig:CSEA_AFSF_fig3}
\end{figure*}

%% file: sec/3_finalcopy.tex
\section{Methods}
In this section, we provide a detailed description of the key components of the proposed SVDC model. The architecture of the model is shown in \cref{fig:backbone_fig2}. We first describe the channel-spatial enhancement attention(CSEA) module for extracting high-frequency regions (in \cref{CSEA}). Next, we introduce the adaptive frequency selective fusion(AFSF) module (in \cref{AFSF}). Finally, we present the design of the cross-window temporal consistency loss function (in \cref{Cross_loss}). The overall loss function is discussed in \cref{tot_loss}.

\subsection{Channel-Spatial Enhancement Attention} \label{CSEA}
Misalignment during the multi-frame feature fusion stage often leads to blending between object edges and the background. To address this issue, inspired by CBAM\cite{wooCBAMConvolutionalBlock2018} and Selective-Stereo\cite{wangSelectiveStereoAdaptiveFrequency2024}, we propose a Channel-Spatial Enhancement Attention (CSEA) module to guide the network to enhance correctly aligned feature while suppressing misaligned one. At the same time, it extracts attention weights to distinguish between high-frequency and low-frequency regions in the features. As shown in \cref{fig:CSEA_AFSF_fig3}. The CSEA module consists of two components: the Channel Enhancement (CE) module and the Spatial Attention (SA) module. The CE module guides the network on what to focus on, while the SA module guides the network on where to focus.

\textbf{Channel Enhancement Module.} 
Given an input feature $F \in \mathbb{R}^{C \times H \times W}$, we can apply average pooling and max pooling along the spatial dimension to obtain $F_{\text{avg}}, F_{\text{max}} \in \mathbb{R}^{C \times 1 \times 1}$. These represent the global average and maximum responses in the $H \times W$ space, helping to better infer which channels should be focused on. Next, we concatenate these features and pass them through two convolutional layers. Finally, we add them together and use sigmoid as the activation function, resulting in channel attention weights in the range of $(0, 1)$, denoted as $A^c \in \mathbb{R}^{C \times 1 \times 1}$. We use these weights to perform element-wise products with the input feature as $A^c \cdot F$, achieving enhancement along the channel dimension. This channel enhancement module adaptively enhances features with higher importance while suppressing less informative features.

\textbf{Spatial Attention Module.} 
Similar to the CE module, the SA module also enhances the features. However, unlike the CE module, The SA module focuses on the areas where attention is required. We apply a pooling operation along the feature dimension now. Then, we concatenate the pooled features to obtain features with a shape of $\mathbb{R}^{2 \times H \times W}$. Then, through a $1 \times 1$ convolutional layer and a sigmoid function, we obtain the final spatial attention weight $A^s \in \mathbb{R}^{1 \times H \times W}$. From the preceding operations, it can be observed that the attention map assigns higher weights to regions requiring high-frequency information, as these features exhibit high values in the fused information, while lower weights are assigned to regions requiring low-frequency information. The CSEA module effectively distinguishes between high-frequency and low-frequency regions, enabling adaptive selection of convolution kernel sizes. This facilitates the recovery of high-frequency details and suppression of abnormal high-frequency information in misaligned regions, thus reducing the impact of misalignment and improving the accuracy and consistency of network predictions.

\subsection{Adaptive Frequency Selective Fusion} \label{AFSF}
To more accurately fuse information across multiple frames and avoid blending issues between objects and the background caused by misalignment, we propose an adaptive frequency selective fusion module. As shown in \cref{fig:CSEA_AFSF_fig3}. This module adaptively applies smaller convolution kernels in high-frequency regions, ensuring that the feature at object edges remains unaffected by the background. For low-frequency regions, such as flat or textureless areas, larger convolution kernels are used to smooth out the impact of high-frequency noise on these regions. 

Specifically, during the multi-frame feature fusion stage, we draw inspiration from the optical flow-guided deformable convolution used in DVSR\cite{sunConsistentDirectTimeFlight2023}. However, while DVSR takes low-resolution dToF depth maps as input, directly applying this method to the fusion of sparse dToF depth maps faces more severe blending issues between objects and the background under misaligned conditions. This is because sparse depth maps provide depth values at pixel-wise coordinates, and the sparse points exhibit larger variations between adjacent frames. In contrast, low-resolution depth maps provide the mean values within patches, resulting in smaller changes.

To address this issue, we utilize the attention maps generated by the CSEA module to distinguish between high-frequency and low-frequency regions. Smaller convolution kernels are applied to process multi-frame features in high-frequency regions, while larger convolution kernels are used for low-frequency regions. This approach ultimately achieves adaptive fusion across regions with different frequencies.

The adaptive fusion stage is defined as follows:
\begin{equation}
F_A^{\text{fused}} = A \cdot F_A^s + (1 - A) \cdot F_A^l
\end{equation}

Where $A$ represents the attention map extracted from the CSA module, which has higher weights in high-frequency regions. $F_A^s$ represents multi-frame features processed by convolutions with small kernels, while $F_A^l$ represents features processed by convolutions with large kernels.

\subsection{Cross-window temporal consistency loss} \label{Cross_loss}
\begin{figure}[t]
  \centering
  \includegraphics[width=1\linewidth]{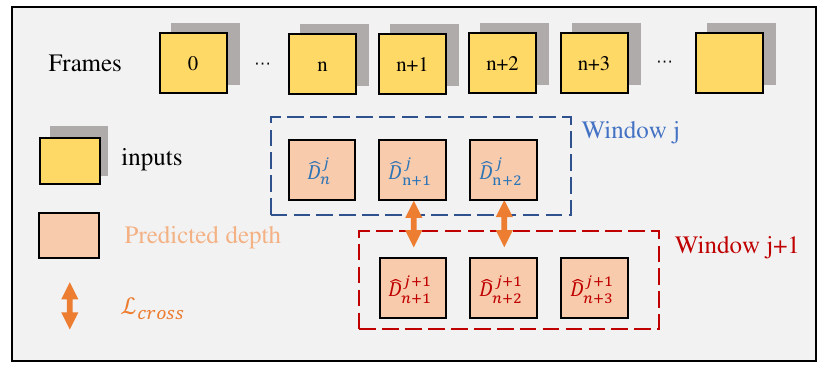}

  \caption{The supervision process of the Cross-Window Temporal Consistency Loss.}
  \label{fig:cross-window-loss}
\end{figure}

To further introduce cross-window information interaction and ensure the consistency of predictions across windows, we propose the cross-window temporal consistency loss $\mathcal{L}_{\text{cross}}$. Taking a window size of 3 frames as an example, its specific illustration is shown in \cref{fig:cross-window-loss}.

During training, each window predicts three consecutive depth maps. The predicted depth map for frame $n$ in window $j$ is denoted as $\hat{D}_n^j$. Due to the feature fusion and the temporal consistency supervision within the window, consistency within the same window is ensured. However, the lack of information interaction across windows makes it difficult to ensure consistent predictions for the same frame across different windows.

Specifically, for the $n+1$ frame predicted in different windows $j$ and $j+1$, the results $\hat{D}_{n+1}^{j}$ and $\hat{D}_{n+1}^{j+1}$ should be identical in the ideal case. However, due to the lack of cross-window consistency constraints, even small differences in the input color and sparse depth maps across different windows may lead to significant variations in the final predictions. We use the Scale-Invariant Loss\cite{eigenDepthMapPrediction2014} (SILoss) to minimize the differences in predictions of the same frame across different windows. By minimizing the variations in prediction results caused by small input differences, the aim is to make the predictions from different windows as consistent as possible and to address the flickering issue in consecutive frames across windows during inference.

The cross-window consistency loss is defined as:
\begin{equation}
\mathcal{L}_{\text{cross}} = \mathcal{L}_{\text{SI}}(\hat{D}_{n+1}^{j}, \hat{D}_{n+1}^{j+1}) + \mathcal{L}_{\text{SI}}(\hat{D}_{n+2}^{j}, \hat{D}_{n+2}^{j+1})
\end{equation}

\subsection{Loss Functions} \label{tot_loss}

For spatial loss that supervises the depth accuracy, we use the scale-invariant loss $\mathcal{L}_{\text{SI}}$\cite{eigenDepthMapPrediction2014}, defined as:
\begin{equation}
\mathcal{L}_{\text{SI}}(\hat{d}_i, d_i) = \alpha \sqrt{\frac{1}{T} \sum_{i} g_i^2 - \frac{\lambda}{T^2} \left(\sum_{i} g_i \right)^2}
\end{equation}
where $g_i = \log \hat{d}_i - \log d_i$, $\hat{d}_i$ represents the estimated depth, $d_i$ represents the ground-truth depth, and $T$ is the total number of valid pixels. In our experiments, we set $\lambda = 0.85$ and $\alpha = 10$ for all our experiments.

For temporal consistency, we adopt the cross-window consistency loss $\mathcal{L}_{\text{cross}}$ and the temporal consistency loss $\mathcal{L}_{\text{OPW}}$ within a window, based on FMNet\cite{wangLessMoreConsistent2022}:
\begin{equation}
\mathcal{L}_{\text{OPW}} = \frac{1}{T} \sum_{j=1}^{T} M^{(j)}_{n \to n-1} \left\| \hat{D}^{(j)}_n - \tilde{D}^{(j)}_{n-1} \right\|_1
\end{equation}

\begin{equation}
M_{n \to n-1}^{(j)} = \exp\left(-\beta \| F_{n} - \tilde{F}_{n-1} \|_2^2 \right)
\end{equation}
where $\tilde{D}_{n-1}$ is the predicted depth $\hat{D}_{n-1}$ warped by the backward optical flow $O_{n \to n-1}$ between input frames $F_n$ and $\tilde{F}_{n-1}$ . In our implementation, we use SpyNet\cite{ranjanOpticalFlowEstimation2017} as our optical flow network. $M_{n \to n-1}^{(j)}$ indicates the occlusion mask calculated based on the warping discrepancy between frame $F_n$ and warped frame $\tilde{F}_{n-1}$. $T$ represents the number of pixels. We set $\beta = 50$ identical to\cite{caoLearningStructureAffinity2021}, 

Finally, the overall loss function $\mathcal{L}_{\text{total}}$ is defined as:
\begin{equation}
\begin{aligned}
\mathcal{L}_{\text{total}} &= \mathcal{L}_{\text{spatial}} + \mathcal{L}_{\text{temporal}} \\
\mathcal{L}_{\text{spatial}} &= \mathcal{L}_{\text{SI}}(\hat{d}_{\text{final}}, d_{\text{gt}}) + \gamma \mathcal{L}_{\text{SI}}(\hat{d}_{\text{coarse}}, d_{\text{gt}}) \\
\mathcal{L}_{\text{temporal}} &= \mathcal{L}_{\text{cross}}(\hat{d}_{\text{coarse}}) + \lambda_{\text{OPW}}^{t} \mathcal{L}_{\text{OPW}}
\end{aligned}
\end{equation}
Where $\hat{d}_{\text{coarse}}$ represents the predicted low-resolution depth, and $\hat{d}_{\text{final}}$ represents the final depth obtained after upsampling.
In our experiments, we set $\gamma = 0.25$, $\lambda_{\text{OPW}}^{t} = 0.125$.

%% file: sec/4_Experiments.tex
\section{Experiments}
\begin{figure*}[!thbp]
  \centering
  \includegraphics[width=1\linewidth]{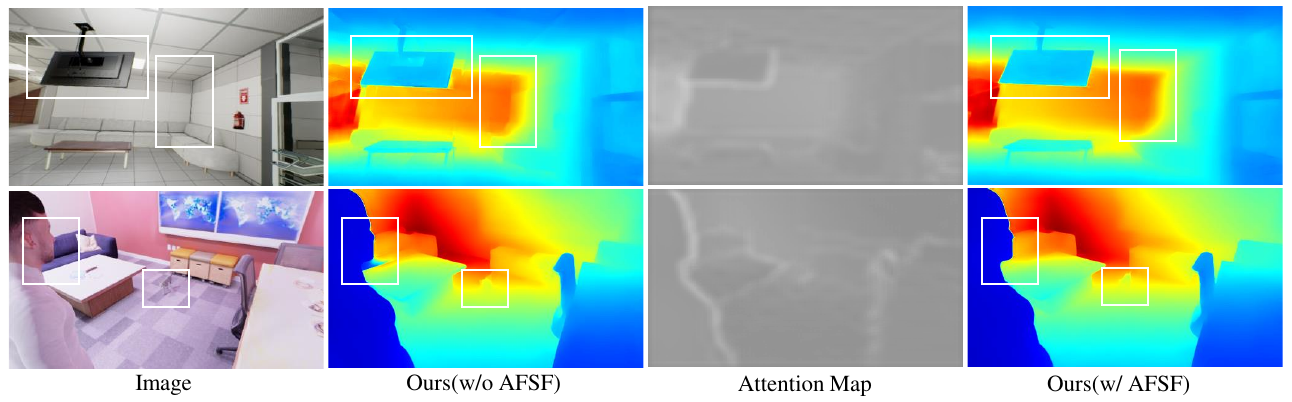}

  \caption{Qualitative results on TartanAir and Dynamic Replica. Row 1: Results on the TartanAir dataset. Row 2: Results on the Dynamic Replica dataset. The third column represents the attention maps extracted by the CSEA module. Our SVDC method outperforms DVDC in both edge prediction and the prediction of smooth regions. }
  \label{fig:AFSF_qualitative_fig5}
\end{figure*}

\begin{table*}[!thbp]
\centering
\begin{tabular}{c|cc|ccccc}
\hline
\multirow{2}{*}{Model} &
  \multirow{2}{*}{CSEA} &
  \multirow{2}{*}{AFSF} &
  \multirow{2}{*}{\begin{tabular}[c]{@{}c@{}}RMSE$\downarrow$\\ (m)\end{tabular}} &
  REL$\downarrow$ &
  \multirow{2}{*}{\begin{tabular}[c]{@{}c@{}}TEPE$\downarrow$\\ (mm)\end{tabular}} &
  OPW$\downarrow$ &
  \multirow{2}{*}{\begin{tabular}[c]{@{}c@{}}Param\\ (M)\end{tabular}} \\
          &            &            &                &                &               &                &      \\ \hline
DVDC      &            &            & 0.183          & 0.030          & 79.8          & 0.166          & 22.7 \\
DVDC+AFSF &            & \checkmark & 0.175          & 0.026          & 73.6          & 0.161          & 22.8 \\
SVDC      & \checkmark & \checkmark & \textbf{0.164} & \textbf{0.024} & \textbf{69.7} & \textbf{0.159} & 22.8 \\ \hline
\end{tabular}
\caption{Ablation study of the effectiveness of CSEA and AFSF modules on the TartanAir dataset.}
\label{tab:abl_CSEA_tab1}
\end{table*}


\textbf{TartanAir}\cite{wangTartanAirDatasetPush2020} is an RGB-D video dataset consisting of a total of 18 scenes, including 15 outdoor scenes and 3 indoor scenes. We only use its easy scene data, which contains 185k pairs of RGB-D images. \textbf{DynamicReplica}\cite{karaevDynamicStereoConsistentDynamic2023} dataset contains 524 synthetic videos of humans and objects performing actions in indoor environments. It consists of 484 training videos, 20 validation videos, and 20 test videos, with a total of 170k pairs of RGB-D images. \textbf{MIPI}\cite{zhuMIPI2023Challenge2023} dataset is a comprehensive dataset of MIPI RGB+ToF depth data, containing 7 indoor scenes with a total of 20k pairs of RGB and depth images. \textbf{DydToF}\cite{sunConsistentDirectTimeFlight2023} dataset proposed in the DVSR paper includes a large amount of dynamic motion information, with a total of 100 scenes and 45k pairs of RGB-D images.

\subsection{Implementation Details}
In our implementation, we train our model based on PyTorch using NVIDIA RTX 3090 GPUs. We generate dToF sparse depth data with a 70° FOV and 30 × 40 sampling points from the ground truth depth map. On this basis, we further introduce barrel distortion, random offsets and rotations, random dropout, and random depth value errors to simulate the noise characteristics of real dToF imaging systems. More details of the dToF sparse depth map can be found in the supplementary materials.

For all experiments, we use the AdamW\cite{loshchilovDecoupledWeightDecay2019} optimizer and clip gradients to the range of [-0.1, 0.1]. We adopt the OnecycleLR scheduler with a maximum learning rate of 3e-4. For the pre-trained optical flow model SpyNet\cite{ranjanOpticalFlowEstimation2017}, we finetune it during training with a learning rate of 3e-5. We use a combination of datasets, including TartanAir, DynamicReplica, MIPI, and DydToF, to train our model. During training, we set window size $T=3$, with a batch size of 6, trained for 200k steps. All images are resized to a resolution of 288 × 512. The training process takes approximately $\sim$3 days on 4$\times$NVIDIA RTX 3090 GPUs.

\subsection{Ablation Study}
In this section, we evaluate our model in different settings to verify our proposed modules in several aspects. 
\begin{figure*}[!htbp]
  \centering
  \includegraphics[width=1\linewidth]{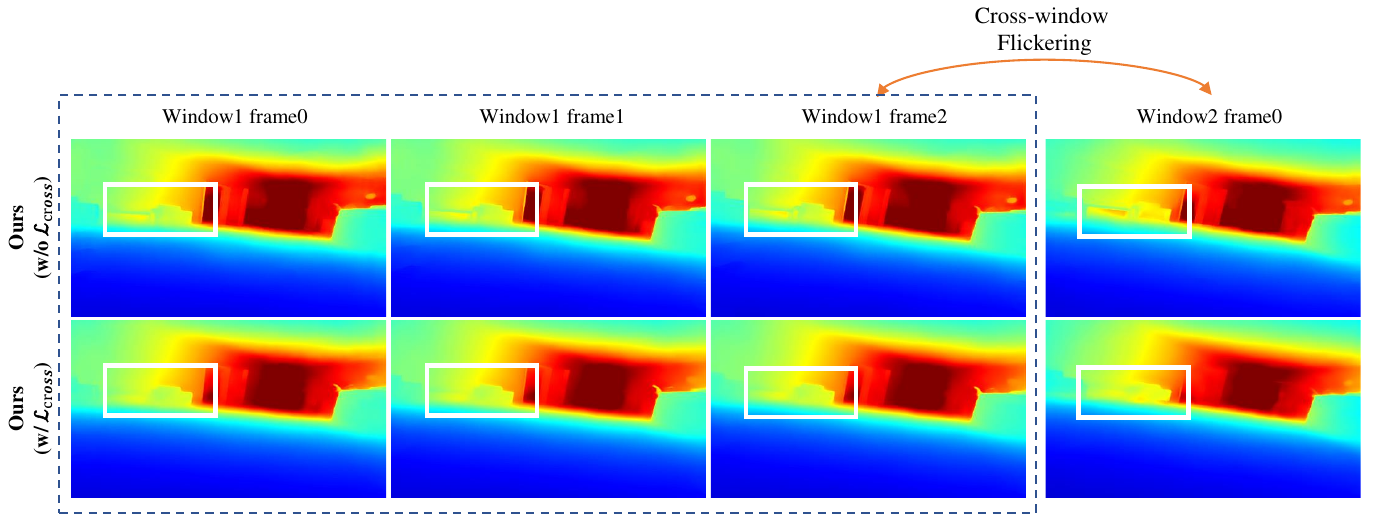}

  \caption{Qualitative results on the TartanAir dataset with the addition of the cross-window consistency loss show that, without window consistency supervision, there is a noticeable flickering phenomenon at the boundaries between frames from different windows. However, after adding the supervision, the flickering issue is alleviated.}
  \label{fig:Cross_window_fig6}
\end{figure*}
\begin{table*}[!thbp]
\centering
\begin{tabular}{c|cc|cc|cc|cccc}
\hline
\multirow{3}{*}{Model} &
  \multirow{3}{*}{\begin{tabular}[c]{@{}c@{}}OPW\\ Loss\end{tabular}} &
  \multirow{3}{*}{\begin{tabular}[c]{@{}c@{}}Cross-Window\\ Loss\end{tabular}} &
  \multicolumn{2}{c|}{Intra-Window} &
  \multicolumn{2}{c|}{Cross-Window} &
  \multicolumn{4}{c}{Average} \\ \cline{4-11} 
 &
   &
   &
  \multirow{2}{*}{\begin{tabular}[c]{@{}c@{}}TEPE$\downarrow$\\ (mm)\end{tabular}} &
  OPW$\downarrow$ &
  \multirow{2}{*}{\begin{tabular}[c]{@{}c@{}}TEPE$\downarrow$\\ (mm)\end{tabular}} &
  OPW$\downarrow$ &
  \multirow{2}{*}{\begin{tabular}[c]{@{}c@{}}TEPE$\downarrow$\\ (mm)\end{tabular}} &
  OPW$\downarrow$ &
  \multirow{2}{*}{\begin{tabular}[c]{@{}c@{}}RMSE$\downarrow$\\ (m)\end{tabular}} &
  REL$\downarrow$ \\
 &
   &
   &
   &
   &
   &
   &
   &
   &
   &
  \multicolumn{1}{l}{} \\ \hline
\multirow{3}{*}{SVDC} &
   &
   &
  17.5 &
  0.145 &
  47.6 &
  0.490 &
  27.5 &
  0.260 &
  0.094 &
  0.024 \\
 &
  \checkmark &
   &
  12.9 &
  0.076 &
  47.1 &
  0.489 &
  24.2 &
  0.212 &
  0.096 &
  0.025 \\
 &
  \checkmark &
  \checkmark &
  \textbf{11.1} &
  \textbf{0.066} &
  \textbf{36.3} &
  \textbf{0.384} &
  \textbf{19.4} &
  \textbf{0.171} &
  \textbf{0.086} &
  \textbf{0.020} \\ \hline
\end{tabular}
\caption{Ablation study of the Cross-window consistency loss on the Dynamic Replica dataset.}
\label{tab:abl_cross_tab3}
\end{table*}


\textbf{Effectiveness of CSEA and AFSF}. 
We tested the results of CSEA and AFSF on the TartanAir dataset, as shown in \cref{tab:abl_CSEA_tab1}. Our AFSF method improves consistency and accuracy by directly adding results from different kernel sizes, even without attention maps, demonstrating the benefit of merging frequency information from varying kernel sizes for network inference. After incorporating CSEA, the network adaptively selects and fuses the results from different kernel sizes based on the Attention map, achieving the best results by adding only 0.1M additional parameters. We present visualizations on the TartanAir\cite{wangTartanAirDatasetPush2020} and Replica\cite{karaevDynamicStereoConsistentDynamic2023} datasets as shown in \cref{fig:AFSF_qualitative_fig5}. With the addition of CSEA and AFSF modules, we achieve improved estimations in both high-frequency edge regions and low-frequency smooth areas. This is due to the network's ability to adaptively preserve high-frequency details while smoothing abnormal noise in low-frequency regions, thus alleviating the impact of optical flow misalignment. The metric results in the edge regions are shown in \cref{tab:edge_result_tab2}. Using the Canny operator to extract the image edges and distinguish between edge and non-edge areas, it is evident that our method achieves optimal performance in both edge and non-edge regions.

\begin{table}[!thbp]
\centering
\begin{tabular}{c|cc|cc}
\hline
\multirow{3}{*}{Model} & \multicolumn{2}{c|}{Edges}      & \multicolumn{2}{c}{Non-Edges}   \\ \cline{2-5} 
 &
  \multirow{2}{*}{\begin{tabular}[c]{@{}c@{}}RMSE$\downarrow$\\ (m)\end{tabular}} &
  REL$\downarrow$ &
  \multirow{2}{*}{\begin{tabular}[c]{@{}c@{}}RMSE$\downarrow$\\ (m)\end{tabular}} &
  REL$\downarrow$ \\
                       &                &                &                &                \\ \hline
DVDC                   & 0.201          & 0.063          & 0.123          & 0.033          \\ \cline{1-1}
SVDC                   & \textbf{0.189} & \textbf{0.058} & \textbf{0.110} & \textbf{0.026} \\ \hline
\end{tabular}
\caption{Quantitative results for different regions on the TartanAir dataset.}
\label{tab:edge_result_tab2}
\end{table}

\begin{table*}[!thbp]
\centering
\begin{tabular}{c|c|cccc|cccc}
\hline
\multirow{3}{*}{Methods} &
  \multirow{3}{*}{\begin{tabular}[c]{@{}c@{}}Params\\ (M)\end{tabular}} &
  \multicolumn{4}{c|}{TartanAir} &
  \multicolumn{4}{c}{Dynamic Replica} \\ \cline{3-10} 
 &
   &
  \multirow{2}{*}{\begin{tabular}[c]{@{}c@{}}RMSE$\downarrow$\\ (m)\end{tabular}} &
  REL$\downarrow$ &
  \multirow{2}{*}{\begin{tabular}[c]{@{}c@{}}TEPE$\downarrow$\\ (mm)\end{tabular}} &
  OPW$\downarrow$ &
  \multirow{2}{*}{\begin{tabular}[c]{@{}c@{}}RMSE$\downarrow$\\ (m)\end{tabular}} &
  REL$\downarrow$ &
  \multirow{2}{*}{\begin{tabular}[c]{@{}c@{}}TEPE$\downarrow$\\ (mm)\end{tabular}} &
  OPW$\downarrow$ \\
 &
   &
   &
  \multicolumn{1}{l}{} &
   &
  \multicolumn{1}{l|}{} &
   &
  \multicolumn{1}{l}{} &
   &
  \multicolumn{1}{l}{} \\ \hline
BPNet &
  89.9 &
  0.337 &
  0.051 &
  159.2 &
  0.256 &
  0.126 &
  0.031 &
  57.0 &
  0.601 \\
CFormer &
  82.5 &
  0.352 &
  0.052 &
  163.4 &
  0.254 &
  0.127 &
  0.030 &
  49.4 &
  0.465 \\
DVDC &
  22.7 &
  0.183 &
  0.030 &
  79.8 &
  0.166 &
  0.095 &
  0.026 &
  24.0 &
  0.208 \\
SVDC &
  22.8 &
  \textbf{0.164} &
  \textbf{0.024} &
  \textbf{69.7} &
  \textbf{0.159} &
  \textbf{0.086} &
  \textbf{0.020} &
  \textbf{19.4} &
  \textbf{0.171} \\ \hline
\end{tabular}
\caption{Quantitative results on the TartanAir and Dynamic Replica datasets. Our multi-frame method achieves the best performance in terms of both accuracy and consistency.}
\label{tab:sota_result_tab4}
\end{table*}

\textbf{Effectiveness of proposed Cross-window Loss}. We evaluate the cross-window loss on the Dynamic Replica\cite{karaevDynamicStereoConsistentDynamic2023} dataset, and the results are shown in \cref{tab:abl_cross_tab3}. We set the window size to 3 and compared the consistency metrics within the window and across windows. Existing methods introduce the OPW loss\cite{wangLessMoreConsistent2022} to align different frames using optical flow, minimizing differences and improving consistency within the window. However, these methods fail to enhance cross-window consistency and slightly reduce accuracy. In contrast, our proposed cross-window loss significantly improves cross-window consistency by constraining the differences in predictions for the same frame across different windows. Furthermore, by constraining the output consistency under slight input variations, we make the feature space representation more compact, improving intra-window consistency and prediction accuracy, and leading to superior results. As shown in \cref{fig:Cross_window_fig6}, qualitative results demonstrate that the lack of cross-window consistency constraints leads to flickering between adjacent frames across different windows.

\subsection{Comparisons with State-of-the-art}
\begin{figure}[t]
  \centering
  \includegraphics[width=1\linewidth]{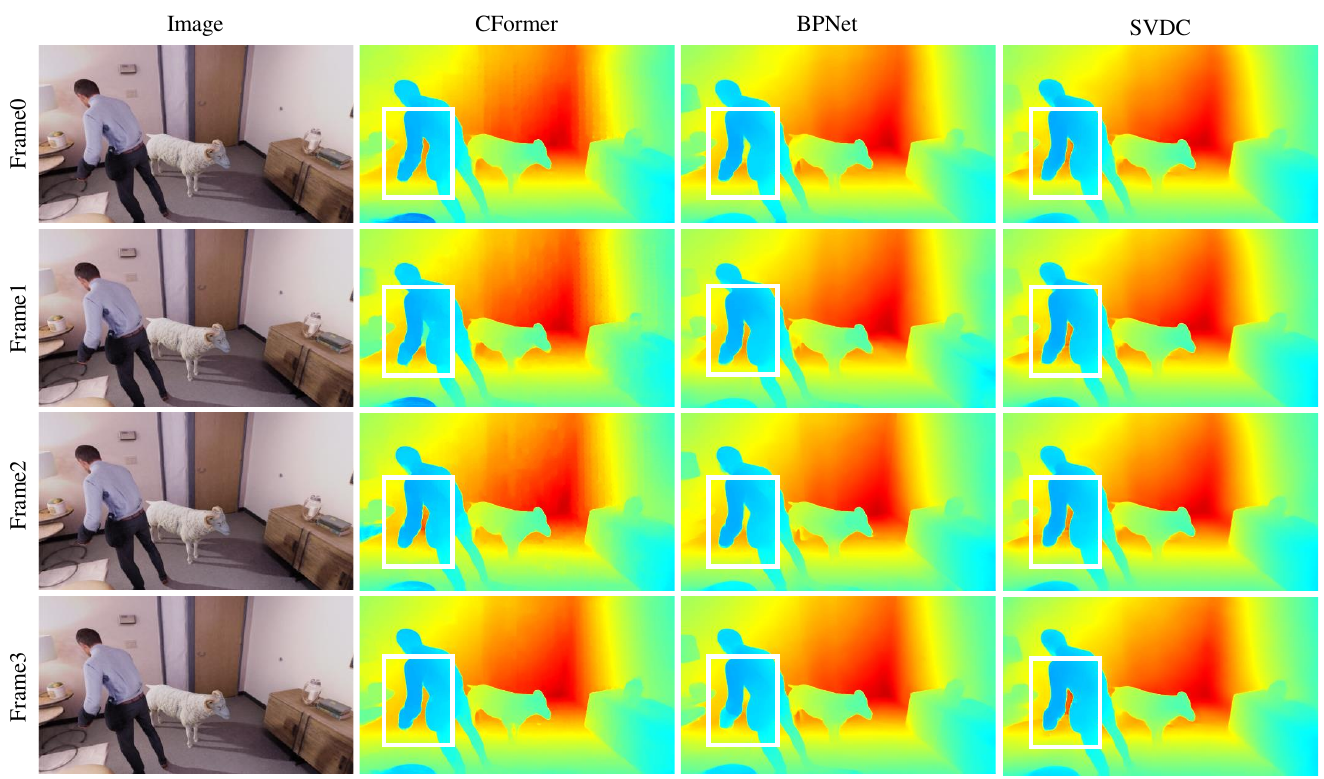}

  \caption{Qualitative comparisons with SOTA methods on the Dynamic Stereo dataset.}
  \label{fig:sota_vis_fig7}
\end{figure}

We evaluate the proposed networks on TartanAir\cite{wangTartanAirDatasetPush2020} and Dynamic Replica\cite{karaevDynamicStereoConsistentDynamic2023} datasets. Since no off-the-shelf algorithms currently utilize dToF sparse depth map for completion, we ensure that the same dToF sparse depth map is used as input to retrain existing state-of-the-art (SOTA) per-frame depth completion networks, such as CFormer\cite{zhangCompletionFormerDepthCompletion2023} and BPNet\cite{tangBilateralPropagationNetwork2024}, under identical training settings to serve as our baselines. We evaluate the methods using four metrics: root mean squared error(RMSE), mean absolute relative error(REL), temporal end-point error (TEPE), and OPW. The results are shown in \cref{tab:sota_result_tab4}. Our multi-frame method achieves the best performance in terms of both accuracy and consistency. We also provide visual comparison results with SOTA methods. As shown in \cref{fig:sota_vis_fig7}, we present visualizations of four consecutive frames. The third and fourth frames span different windows, where our method exhibits stable performance in predicting the person in the images, while other methods show noticeable flickering.

\textbf{TartanAir}
We evaluate using two scenes from the TartanAir\cite{wangTartanAirDatasetPush2020} dataset, with 300 frames in each scene. Our multi-frame fusion method achieves the best results, outperforming current state-of-the-art (SOTA) per-frame methods, with particularly significant improvements in consistency. In contrast, existing per-frame methods perform suboptimally due to the assumption that sparse depth maps are accurate. Networks like CSPN\cite{chengCSPNLearningContext2020}, for example, perform iterative optimization based on sparse depth and propagate information from surrounding points. However, the dToF sparse depth maps are highly noisy, which can cause the noise to propagate and lead to suboptimal results. By fusing information across multiple frames, we successfully help the completion of dToF sparse depth map.

\textbf{Dynamic Replica}
The Dynamic Replica\cite{karaevDynamicStereoConsistentDynamic2023} is an indoor dataset that contains a large number of moving objects. As a result, the temporal consistency error in single-frame network estimates tends to be relatively high. However, the method we propose adaptively fuses multi-frame features, and with the addition of a temporal consistency constraint, it shows a significant improvement in temporal consistency compared to the single-frame method. Moreover, Our method also achieves the best accuracy.

%% file: sec/5_Conclusion.tex
\section{Conclusion}
In this paper, we propose a multi-frame approach for dToF depth completion to address sparse and noisy depth maps in mobile devices. By combining a lightweight optical flow model with convolution, and introducing the Adaptive Frequency Selective Fusion (AFSF) and Channel-Spatial Enhancement Attention (CSEA) modules, our method improves depth prediction accuracy and preserves object boundaries. Extensive experiments on the TartanAir and Dynamic Replica datasets demonstrate that our approach outperforms existing methods, achieving superior performance with fewer parameters.

However, our methods still face some challenges. Firstly, our method relies on a pre-trained optical flow model, which may struggle in conditions where optical flow estimation is particularly challenging, such as low-light environments or scenes with large motion. Secondly, the sparse and noisy dToF depth map leads to inefficient information propagation and potential errors. We could explore densifying dToF features during the preprocessing stage to help the network learn more effectively. Finally, exploring the use of cross-attention for implicit multi-frame fusion and learning motion representations is a promising direction, although the computational overhead needs to be carefully considered.

%% file: sec/X_suppl.tex
\clearpage
\setcounter{page}{1}
\maketitlesupplementary
\appendix
This supplementary material provides additional information to complement the main paper. It contains the following sections:\\
\begin{itemize}
    \item More experimental results in Sec. \ref{sec:Experimentresults}.\\
    
    \item More implementation details in Sec. \ref{sec:implementationdetails}.\\
    
    \item Network architecture details in Sec. \ref{sec:Network}.\\
    
    \item More qualitative results in Sec. \ref{sec:Qualitative}.
\end{itemize}

\section{More Experimental Results}
\label{sec:Experimentresults}

In this section, we present additional experimental results.
\subsection{Ablation Study on Kernel Sizes}
We conducted an ablation study on the kernel size within the Adaptive Frequency Selective Fusion (AFSF) module. The detailed results are shown in \cref{tab:kernel_sizes}. Considering both accuracy and temporal consistency, we ultimately selected the combination of $1 \times 1$  and $3 \times 3$ convolutional kernels as our experimental configuration. 

\begin{table}[!htbp]
\centering
\resizebox{\columnwidth}{!}{%
\begin{tabular}{cl|ccc|ccc}
\hline
\multicolumn{2}{c|}{\multirow{2}{*}{\begin{tabular}[c]{@{}c@{}}Kernel\\ Sizes\end{tabular}}} &
  \multicolumn{3}{c|}{TartanAir\cite{wangTartanAirDatasetPush2020}} &
  \multicolumn{3}{c}{Dynamic Replica\cite{karaevDynamicStereoConsistentDynamic2023}} \\ \cline{3-8} 
\multicolumn{2}{c|}{}        & RMSE(m)        & REL            & OPW   & RMSE(m)        & REL            & OPW   \\ \hline
\multicolumn{2}{c|}{1$\times$1 + 5$\times$5} & 0.173          & 0.025          & 0.163 & \textbf{0.082} & \textbf{0.020} & 0.175 \\
\multicolumn{2}{c|}{3$\times$3 + 5$\times$5} & \textbf{0.164} & \textbf{0.024} & 0.172 & 0.084          & 0.021          & 0.201 \\
\multicolumn{2}{c|}{1$\times$1 + 3$\times$3} &
  \textbf{0.164} &
  \textbf{0.024} &
  \textbf{0.159} &
  0.086 &
  \textbf{0.020} &
  \textbf{0.171} \\ \hline
\end{tabular}%
}
\caption{Comparison of different kernel sizes on TartanAir and Dynamic Replica datasets.}
\label{tab:kernel_sizes}
\end{table}

\subsection{Computational Cost of Methods}
We evaluated the parameter count and computational cost of different completion methods, as detailed in \cref{tab:computecost}. It can be observed that our proposed baseline model for multi-frame fusion, DVDC, achieves the smallest parameter count and FLOPs. Building on this baseline, the SVDC model, which incorporates CSEA and AFSF, increases the parameter count by only 0.1M and the FLOPs by 3.4 GFLOPs, demonstrating the lightweight characteristics of our proposed design.

\begin{table}[!htbp]
\centering
\begin{tabular}{ccccc}
\hline
           & CFormer & BPNet & DVDC & SVDC \\ \hline
FLOPs (G)  & 184.1  & 247.9 & 48.2 & 51.6 \\
Params (M) & 82.5    & 89.9  & 22.7 & 22.8 \\ \hline
\end{tabular}
\caption{Comparison of computational cost and the parameters.}
\label{tab:computecost}
\end{table}

\subsection{More Quantitative Comparisons}
In the accuracy comparison between our method and the SOTA methods, only RMSE and REL are used. Additional results on the TartanAir and Dynamic Replica datasets are shown in \cref{tab:more_tt} and \cref{tab:more_ds}.


\begin{table}[!htbp]
\centering
\begin{tabular}{c|ccccc}
\hline
\multirow{3}{*}{Methods} & \multicolumn{5}{c}{TartanAir}                  \\ \cline{2-6} 
     & \multirow{2}{*}{\begin{tabular}[c]{@{}c@{}}RMSE$\downarrow$\\ (m)\end{tabular}} & REL$\downarrow$ & $\delta_1\uparrow$ & $\delta_2\uparrow$ & $\delta_3\uparrow$ \\
                         &       &       &       &       &                \\ \hline
BPNet                    & 0.337 & 0.051 & 0.965 & 0.976 & 0.983          \\
CFormer                  & 0.352 & 0.052 & 0.963 & 0.975 & 0.982          \\
DVDC                     & 0.183 & 0.030 & 0.994 & 0.998 & \textbf{0.999} \\
SVDC & \textbf{0.164}                                                                  & \textbf{0.024}  & \textbf{0.995}     & \textbf{0.999}     & \textbf{0.999}     \\ \hline
\end{tabular}
\caption{Quantitative results on the TartanAir dataset.}
\label{tab:more_tt}
\end{table}

\begin{table}[!htbp]
\centering
\begin{tabular}{c|ccccc}
\hline
\multirow{3}{*}{Methods} & \multicolumn{5}{c}{Dynamic Replica}                                                \\ \cline{2-6} 
 & \multirow{2}{*}{\begin{tabular}[c]{@{}c@{}}RMSE$\downarrow$\\ (m)\end{tabular}} & REL$\downarrow$ & $\delta_1\uparrow$ & $\delta_2\uparrow$ & $\delta_3\uparrow$ \\
                         &                &                &                &                &                \\ \hline
BPNet                    & 0.126          & 0.031          & 0.987          & 0.993          & 0.995          \\
CFormer                  & 0.127          & 0.030          & 0.986          & 0.993          & 0.995          \\
DVDC                     & 0.095          & 0.026          & 0.993          & 0.997          & \textbf{0.998} \\
SVDC                     & \textbf{0.086} & \textbf{0.020} & \textbf{0.994} & \textbf{0.998} & \textbf{0.998} \\ \hline
\end{tabular}
\caption{Quantitative results on the Dynamic Replica dataset.}
\label{tab:more_ds}
\end{table}

\section{More Implementation Details}
\label{sec:implementationdetails}
\subsection{Sparse dToF Data}
\begin{figure}[!htbp]
  \centering
  \includegraphics[width=1.0\linewidth]{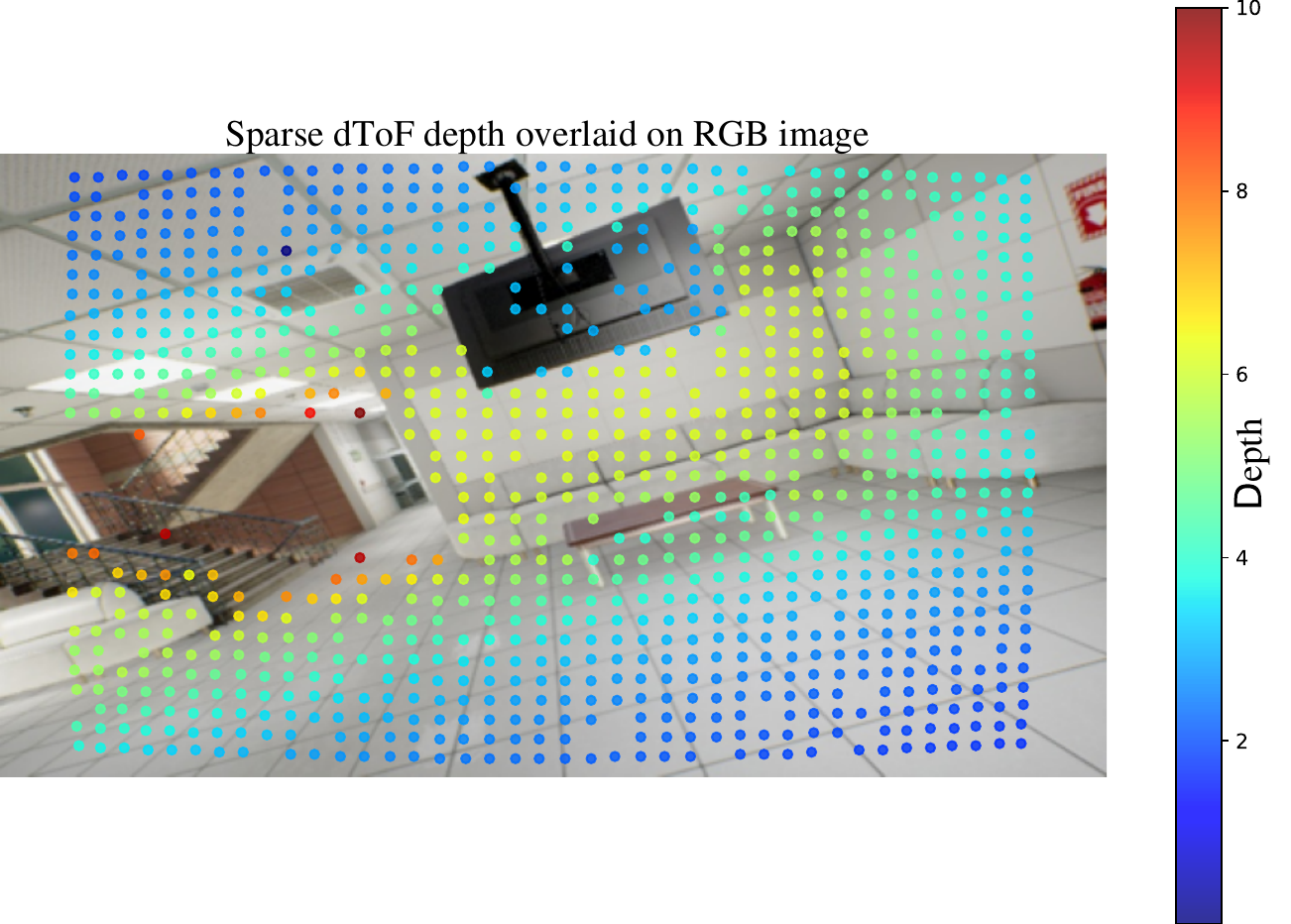}

  \caption{Sparse dToF depth on RGB image}
  \label{fig:sup_dtof}
\end{figure}

When simulating actual dToF data from ground truth depth, several steps are taken to make the simulated sparse dToF depth closely resemble those collected by real-world devices. The field of view (FOV) is set to 70°, and a uniform sampling of $30\times40$ pixels is applied. Barrel distortion is introduced, along with global rotation and translation transformations. Points with low reflectance are dropped based on their RGB values. Random noise and dropout are also added to the data. The visualized results of the simulated sparse dToF depth are shown in \cref{fig:sup_dtof}.

These perturbations significantly degrade the quality of the sparse dToF depth. The RMSE and REL of the valid depth points returned by the dToF simulation are summarized in \cref{tab:sup_dtof_table}. On the TartanAir dataset, the REL is 0.060, and the RMSE is 0.494, while on the Dynamic Replica dataset, the REL is 0.058, and the RMSE is 0.292.

\begin{table}[!htbp]
\begin{tabular}{cl|cc|cc}
\hline
\multicolumn{2}{c|}{Input data}                                                                   & \multicolumn{2}{c|}{TartanAir} & \multicolumn{2}{c}{Dynamic Replica} \\ \hline
\multicolumn{2}{c|}{\multirow{2}{*}{\begin{tabular}[c]{@{}c@{}}Sparse\\ dToF depth\end{tabular}}} & RMSE(m)         & REL          & RMSE(m)           & REL             \\ \cline{3-6} 
\multicolumn{2}{c|}{}                                                                             & 0.494           & 0.060        & 0.292             & 0.058           \\ \hline
\end{tabular}
\caption{Sparse dToF depth metrics}
\label{tab:sup_dtof_table}
\end{table}

\begin{figure*}[!htbp]
  \centering
  \includegraphics[width=1.0\linewidth]{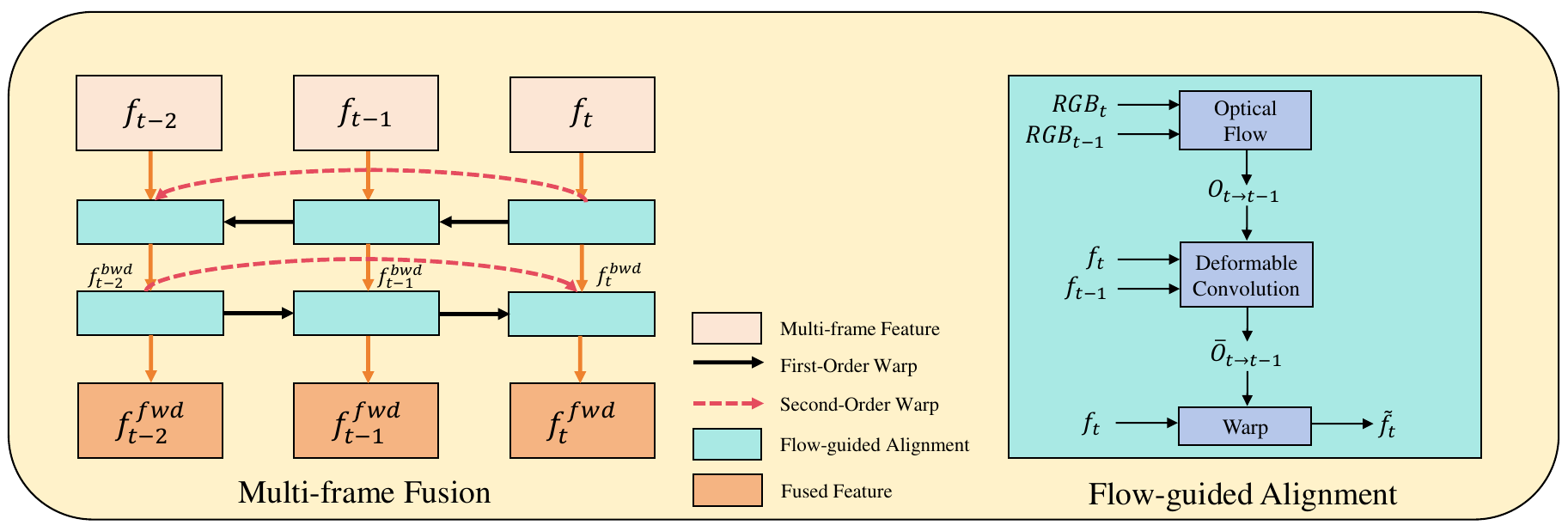}

  \caption{Multi-frame fusion network details}
  \label{fig:sup_multiframe}
\end{figure*}


\subsection{Definition of Evaluation Metrics}
We provide the definitions of the metrics used during our testing. The temporal consistency metric OPW\cite{wangLessMoreConsistent2022} has already been mentioned in the main text of the paper. Here, we supplement it with detailed explanations of the accuracy metrics RMSE, REL, and Accuracy with threshold t, as well as the temporal consistency metric TEPE\cite{sunConsistentDirectTimeFlight2023}.

\begin{itemize}
    \item \textbf{Accuracy Metrics}
\end{itemize}

\textbf{Root Mean Square Error (RMSE):}
\[
\text{RMSE} = \sqrt{\frac{1}{N} \sum_{i=1}^{N} (\hat{d}_i - d_i)^2}
\]
where $\hat{d}_i$ represents the predicted depth, $d_i$ represents the ground truth depth, and $N$ is the number of valid pixels.

\textbf{Mean Absolute Relative Error (REL):}
\[
\text{REL} = \frac{1}{N} \sum_{i=1}^{N} \frac{\lvert \hat{d}_i - d_i \rvert}{d_i}
\]
where $\hat{d}_i$ represents the predicted depth, $d_i$ represents the ground truth depth, and $N$ is the number of valid pixels.

\textbf{Accuracy with threshold t:} Percentage of $d_i$ such that 
\[
\max\left(\frac{\hat{d}_i}{d_i}, \frac{d_i}{\hat{d}_i}\right) = \delta < t ,\quad t \in \{1.25, 1.25^2, 1.25^3\},
\]
where $\hat{d}_i$ and $d_i$ are the predicted depth and ground truth depth of pixel $i$.

\begin{itemize}
    \item \textbf{Temporal Consistency Metric}
\end{itemize}

\textbf{Temporal End-Point Error (TEPE):}
\[
\text{TEPE} = \|\big(\mathcal{W}(d_i) - d_{i+1}\big) - \big(\mathcal{W}(\hat{d}_i) - \hat{d}_{i+1}\big)\|_1
\]
where $\mathcal{W}(\cdot)$ represents the optical flow warping operation from frame $i$ to frame $i+1$. We use the optical flow predicted by the GMFlow\cite{xuGMFlowLearningOptical2022} to perform this warping.

\section{Network Architecture Details}
\label{sec:Network}
\subsection{Multi-frame Fusion}

The multi-frame fusion network architecture is shown in \cref{fig:sup_multiframe}. Multi-frame features are aligned using a flow-guided network and then sent to a bidirectional propagation module, where feature fusion is performed using a Res-block\cite{heDeepResidualLearning2016}. Taking the alignment of features between the \(t\)-th and \((t-1)\)-th frames as an example, the optical flow-guided alignment network first inputs \(RGB_t\) and \(RGB_{t-1}\) into the pre-trained optical flow model SpyNet\cite{ranjanOpticalFlowEstimation2017} to obtain the coarse optical flow \(O_{t \to t-1}\). Then, \(O_{t \to t-1}\) and features \(f_t\), \(f_{t-1}\) are concatenated, sent into a deformable convolutional network\cite{daiDeformableConvolutionalNetworks2017} to derive the refined optical flow \(\overline{O}_{t \to t-1}\). Due to the diversity of the deformable convolution network, we can obtain 8 different offsets to flexibly extract features near the corresponding pixels. Finally, we warp the feature \(f_t\) with the fine optical flow \(\overline{O}_{t \to t-1}\), obtaining the feature \(\tilde{f}_{t}\), aligned with \(f_{t-1}\).
\begin{align}
O_{t \to t-1} &= \mathit{SpyNet}(RGB_t, RGB_{t-1}) \\
\overline{O}_{t \to t-1} &= \mathit{DCN}(concat(f_t, f_{t-1}), O_{t \to t-1}) \\
\tilde{f}_t &= \mathcal{W}(f_t, \overline{O}_{t \to t-1})
\end{align}

\subsection{DepthHead}

We employ the method proposed in AdaBins\cite{bhatAdaBinsDepthEstimation2021}, replacing its miniViT module with a lightweight convolutional module as our depth head, which maps the feature representations to the depth. Unlike directly regressing depth, we predict the depth as a linear combination of different depth bins. Specifically, for each image, we predict its bin-width vector \(b\), which is used to derive the depth bin centers \(c(b)\). For each pixel, we predict its probabilities \(p\) of belonging to different bins. Assuming the depth range is divided into \(N\) different bins, the final predicted depth \(\hat{d}\) for each pixel can be expressed as follows:

\begin{equation}
\hat{d} = \sum_{k=1}^{N} c(b_k) p_k
\end{equation}

\section{More Qualitative Results}

\begin{figure}[!htbp]
  \centering
  \includegraphics[width=1.0\linewidth]{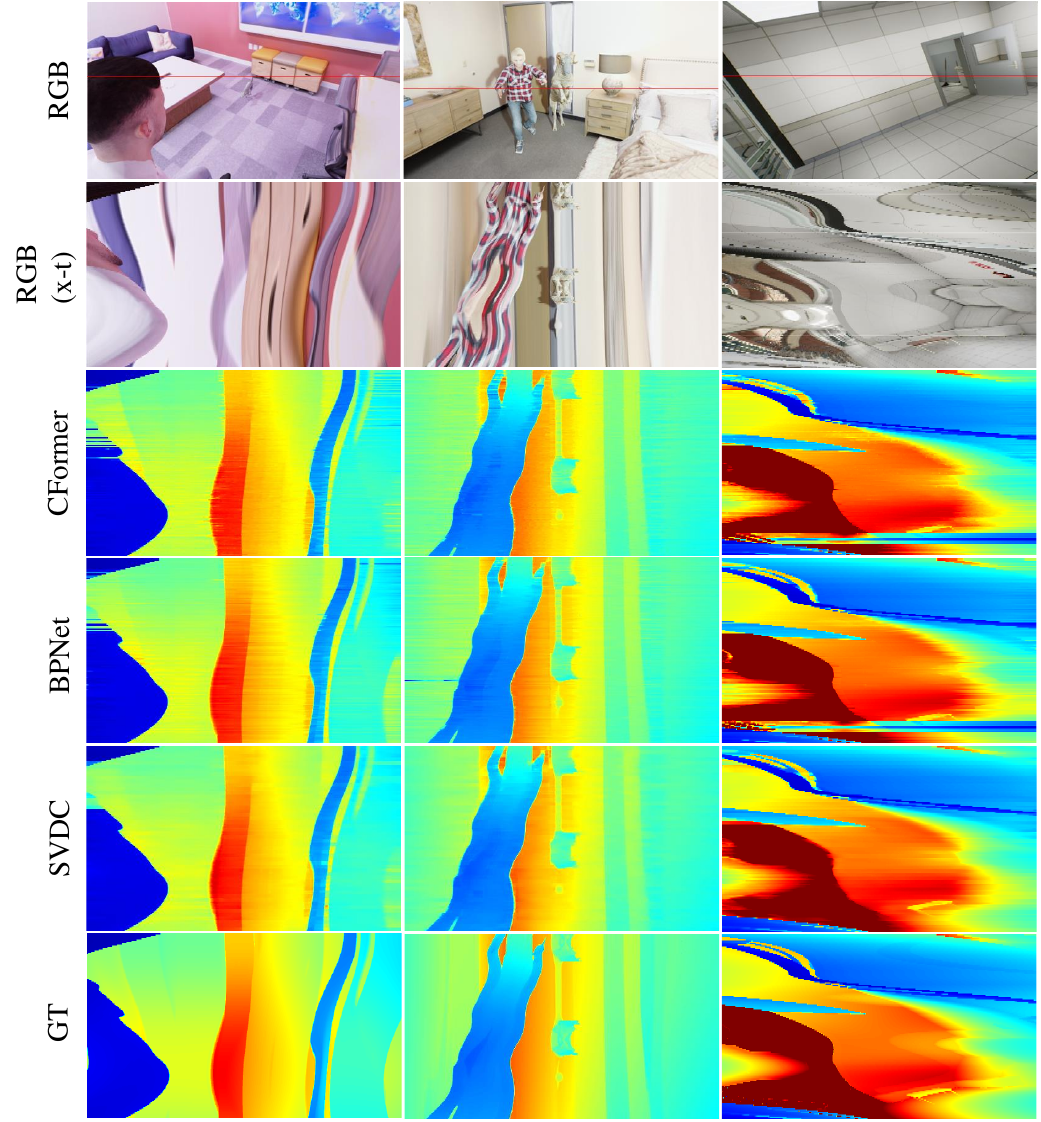}

  \caption{Qualitative results of scanline slice over time}
  \label{fig:sup_x-tfig}
\end{figure}

\label{sec:Qualitative}
In this section, we provide additional visual comparisons on the TartanAir and Dynamic Replica datasets. We plotted scanline slice over time to illustrate the temporal consistency of different methods. Moreover, we also present comparisons of the predictions made by various methods\cite{tangBilateralPropagationNetwork2024,zhangCompletionFormerDepthCompletion2023} in object edges(high-frequency) and smooth regions(low-frequency), highlighting their differences.

In \cref{fig:sup_x-tfig}, we present scanline slice over time, where the first row corresponds to RGB images and the second row represents the scanline patterns over time. Fewer zigzag patterns indicate better temporal consistency. Compared to other methods, our approach demonstrates fewer zigzag patterns, showcasing superior temporal consistency.

\begin{figure*}[!htbp]
  \centering
  \includegraphics[width=1.0\linewidth]{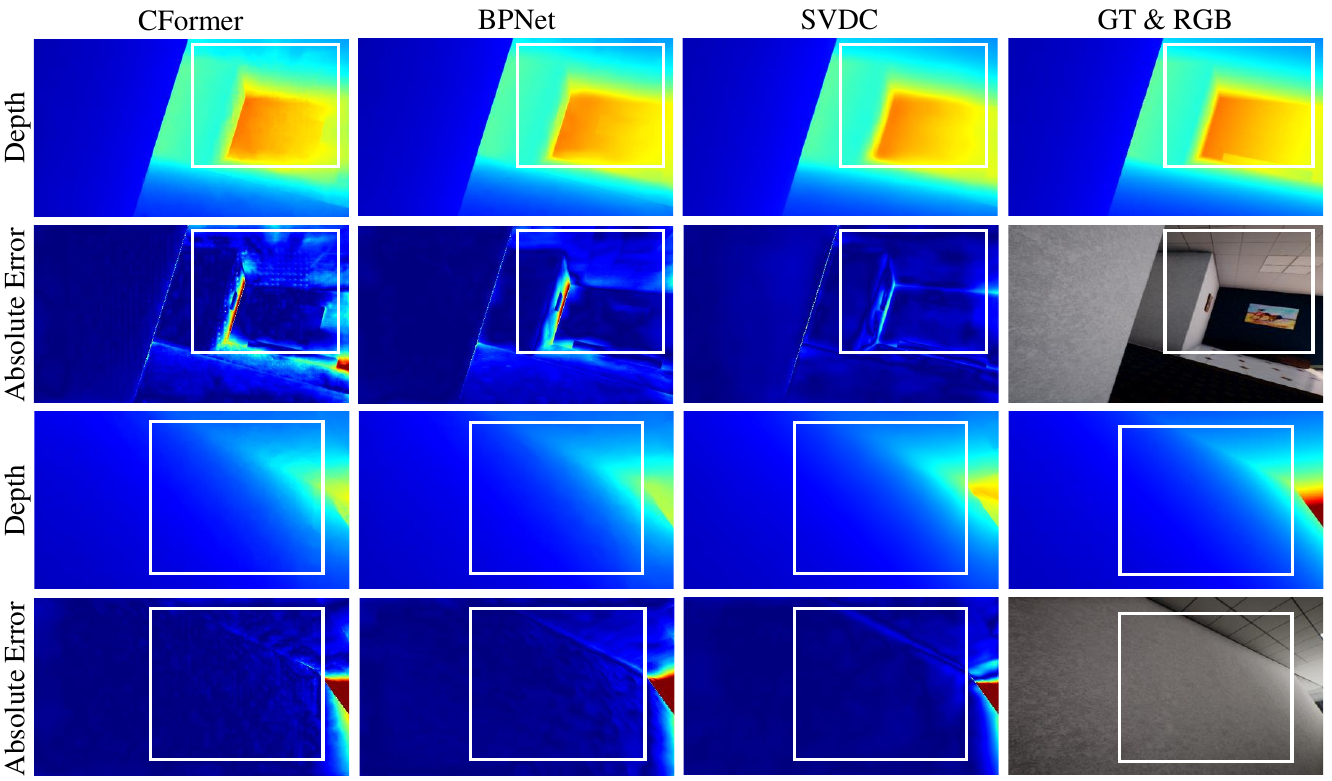}

  \caption{More qualitative results on the TartanAir dataset}
  \label{fig:sup_TartanAir_error}
\end{figure*}

In \cref{fig:sup_TartanAir_error}, we display qualitative results on the TartanAir dataset. It can be observed that our SVDC method achieves smoother estimations in low-frequency regions, demonstrating the effectiveness of our frequency-selective fusion strategy in suppressing high-frequency noise in low-frequency areas.

\begin{figure*}[!htbp]
  \centering
  \includegraphics[width=1.0\linewidth]{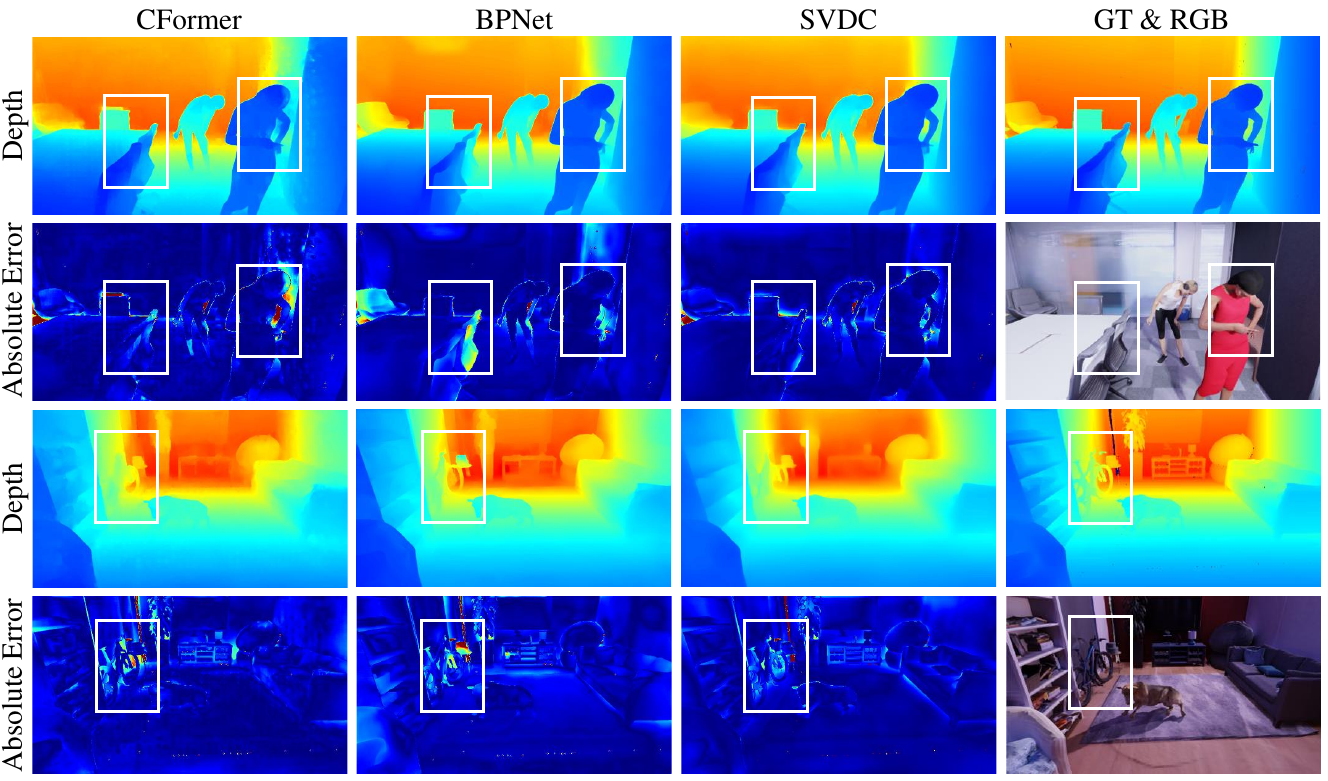}

  \caption{More qualitative results on the Dynamic Replica dataset}
  \label{fig:sup_replica_error}
\end{figure*}

In \cref{fig:sup_replica_error}, we present qualitative results on the Dynamic Replica dataset. The results show that our SVDC method achieves more accurate estimations in high-frequency regions, highlighting its capability to preserve high-frequency details effectively.

